\documentclass{article}
\usepackage{ijcai25}
\usepackage{bbding}
\usepackage{float}
\usepackage{times}
\usepackage{soul}
\usepackage{url}
\usepackage[hidelinks]{hyperref}
\usepackage[utf8]{inputenc}
\usepackage[small]{caption}
\usepackage[pdftex]{graphicx}
\usepackage{amsmath}
\usepackage{amsthm}
\usepackage{booktabs}
\usepackage{algorithm}
\usepackage{algorithmic}
\usepackage[switch]{lineno}
\usepackage{multirow} 
\usepackage{xcolor}
\usepackage{dsfont}

\title{AttentionDrag: Exploiting Latent Correlation Knowledge in Pre-trained Diffusion Models for Image Editing}

\author{
Biao Yang$^1$\and
Muqi Huang$^2$\and
Yuhui Zhang$^1$ \and
Yun Xiong$^1$\thanks{Corresponding authors} \and
Kun Zhou$^2$\footnotemark[1] \and
Xi Chen$^1$\and
Shiyang Zhou$^1$\and
Huishuai Bao$^1$\and
Chuan Li$^2$\and
Feng Shi$^2$\And
Hualei Liu$^2$\\
\affiliations
$^1$Shanghai Key Laboratory of Data Science, School of Computer Science, Fudan University\\
$^2$Rajax Network Technology (ele.me), Alibaba Group\\
\emails
\{biaoyang22, x\_chen21, yuhuizhang23, hsbao22, shiyangzhou22\}@m.fudan.edu.cn,
yunx@fudan.edu.cn,
\{huangmuqi.hmq, kun.zhouk, lc357677, sam.sf\}@alibaba-inc.com,
hualeiliu.taobao@outlook.com
}

\begin{document}

\maketitle

\begin{abstract}
    Traditional point-based image editing methods rely on iterative latent optimization or geometric transformations, which are either inefficient in their processing or fail to capture the semantic relationships within the image. These methods often overlook the powerful yet underutilized image editing capabilities inherent in pre-trained diffusion models. In this work, we propose a novel one-step point-based image editing method, named \textbf{AttentionDrag}, which leverages the inherent latent knowledge and feature correlations within pre-trained diffusion models for image editing tasks. This framework enables semantic consistency and high-quality manipulation without the need for extensive re-optimization or retraining. Specifically, we reutilize the latent correlations knowledge learned by the self-attention mechanism in the U-Net module during the DDIM inversion process to automatically identify and adjust relevant image regions, ensuring semantic validity and consistency. Additionally, AttentionDrag adaptively generates masks to guide the editing process, enabling precise and context-aware modifications with friendly interaction. Our results demonstrate a performance that surpasses most state-of-the-art methods with significantly faster speeds, showing a more efficient and semantically coherent solution for point-based image editing tasks. Code is released at: https://github.com/GPlaying/AttentionDrag.
\end{abstract}

\section{Introduction}

\begin{figure}
    \centering
    \includegraphics[width=1\linewidth]{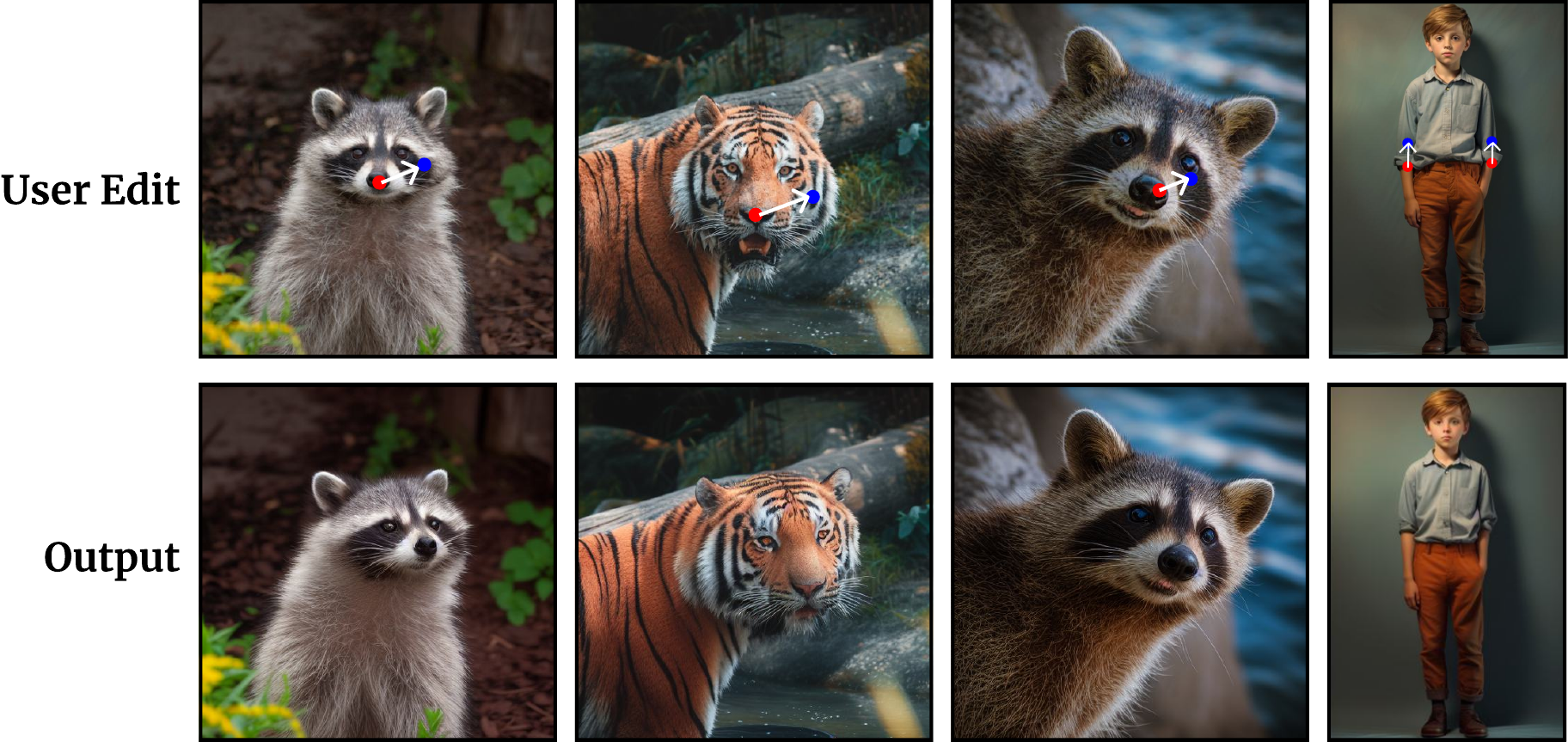}
    \caption{We propose the \textbf{AttentionDrag} algorithm, which leverages the latent correlations knowledge inherent in pre-trained diffusion models to adaptively generate mask regions, achieving semantic consistency and efficient image editing.}
    \label{fig:intro}
\end{figure}

Generative image manipulation has made significant strides with the advent of point-based editing techniques ~\cite{ho2020denoising,rombach2022high,song2020denoising,10681101,CZ2024}. These pointed-based editing methods allow users to intuitively modify images by dragging points to new locations, enabling precise spatial control over specific regions while maintaining the semantic coherence of the image. Point-based editing techniques are particularly valuable for fine-grained image adjustments in generative models, offering more direct and controlled manipulation compared to traditional text-based methods ~\cite{mokady2023null,cho2024noise,ju2023direct,xu2023inversion}.

However, existing point-based image editing methods often face two main limitations. First, traditional point-based methods often rely on iterative optimization techniques, either through motion-based approaches ~\cite{cui2024stabledrag,pan2023drag,shi2024dragdiffusion,ling2023freedrag,liu2024drag} or gradient-based methods~\cite{mou2023dragondiffusion,mou2024diffeditor}. These approaches, such as DragDiffusion~\cite{shi2024dragdiffusion} and GoodDrag~\cite{zhang2024gooddrag}, typically require several steps to refine latent representations, aiming to align manipulated points with desired semantic outcomes. Second, the iterative nature of these methods can introduce instability and fail to preserve semantic consistency across the entire image, especially when the editing involves complex image regions or objects with intricate relationships. To address these issues, some recent methods like FastDrag~\cite{zhao2024fastdrag} attempt to simplify the process by proposing a one-step, geometry-based solution for latent space manipulation. While these methods accelerate the editing process, they still rely on user input, such as manual mask, which reduces usability and flexibility in interactive settings. Additionally, they ignore the intrinsic semantic correlations within the image, thereby hindering their ability to produce semantically consistent and contextually accurate edits.

We argue that existing methods have not fully exploited the latent correlations inherent in pre-trained diffusion models. Pre-trained diffusion models have learned latent knowledge and feature correlations through the self-attention mechanism in their U-Net~\cite{U-Net}. These correlations encompass semantic relationships between different regions of the latent space as well as dependencies between various semantic features (such as objects, textures, and styles) within an image. When leveraged appropriately, these latent correlations knowledge can be directly applied to image editing tasks, achieving superior results without the need for additional training or fine-tuning.

In this work, we introduce \textbf{AttentionDrag}, a novel one-step, training-free method for point-based image editing that fully harnesses the latent correlation knowledge embedded in pre-trained diffusion models. Unlike previous methods, AttentionDrag leverages the self-attention mechanism from the Denoising Diffusion Implicit Models (DDIM) inversion process in the U-Net module to intelligently identify and adjust relevant image regions. This method comprises three key components: First, \textbf{Semantic-based Element Movement} calculates the displacement of image elements in latent space based on their semantic correlations with nearby regions. Next, \textbf{Automatic Mask Generation} leverages the semantic relationships between the handle point and surrounding image elements to automatically identify the editable region. Finally, \textbf{Semantic-based Interpolation} exploits the semantic relationships between the blank regions and their surroundings to transfer relevant semantic information, effectively filling the blank regions created during the editing process. This overall framework ensures that both local and global semantics are preserved, enabling more precise and semantically coherent image manipulations. By exploiting the latent feature space within pre-trained diffusion models, AttentionDrag performs high-quality edits without the need for iterative optimization or extensive retraining, offering a simple, efficient, and powerful alternative to existing techniques.

\textbf{The contributions are as follows:} (1) we demonstrate that pre-trained diffusion models' latent correlation knowledge can be deeply explored and effectively reutilized for image editing tasks; (2) we propose a fast and powerful one-step point-based editing method AttentionDrag, which uses latent feature correlation to achieve semantic consistency; (3) we effectively leverage latent correlation knowledge to identify the editing regions and adaptively generate masks, simplifying the editing process while ensuring semantic consistency and high-quality results; and (4) we conduct comprehensive qualitative and quantitative evaluations to demonstrate the versatility and generalizability of AttentionDrag, showcasing its superior performance in both speed and editing quality compared to existing methods.

\section{Related Work}
\subsection{Text-based Image Editing}
Text-based image editing facilitates a accessible and intuitive editing process, allowing users to execute modifications through natural language interactions. As diffusion models have achieved significant success in image generation, numerous text-based image editing methods have been proposed~\cite{kim2022diffusionclip,brooks2023instructpix2pix,hertz2022prompt} based on diffusion models. These methods edit images by manipulating text prompts. DiffusionCLIP ~\cite{kim2022diffusionclip}was the first to achieve text-guided image manipulation using diffusion models combined with Contrastive Language–Image Pre-training (CLIP) loss. Subsequently, some methods~\cite{brooks2023instructpix2pix,yildirim2023inst,geng2024instructdiffusion} have employed text instructions for training. For example, InstructPix2Pix~\cite{brooks2023instructpix2pix} utilizes a pre-trained large language model with a text-to-image model to generate a large dataset of image editing examples, which is then used to train a conditional diffusion model. Moreover, several methods~\cite{tumanyan2023plug,hertz2022prompt,manukyan2023hd}have achieved training and fine-tuning free editing by modifying attention mechanisms. For instance, Prompt-to-Prompt ~\cite{hertz2022prompt}firstly enables intuitive image editing through textual prompts without the refinement of diffusion models. However, as many editing information are difficult to convey through natural language description, text-based methods lacks in space precision and explicit control, which is drag-based methods specialize in.

\subsection{Point-based Image editing}
Point-based image editing aims to achieve precise spatial control based on user's drag instructions. These methods can be divided into two categories: multi-step optimization methods ~\cite{pan2023drag,shi2024dragdiffusion,zhao2024fastdrag,hou2024easydrag,cho2024noise,zhang2024gooddrag,mou2023dragondiffusion,mou2024diffeditor} and one-step optimization methods ~\cite{zhao2024fastdrag}.

\paragraph{Multi-step Optimization.} Due to the correspondence between the latent space and the pixel space, methods based on latent optimization adopt a step-by-step optimization in latent space to achieve image editing. DragGAN~\cite{pan2023drag} was the first to propose drag-based image editing using Generative Adversarial Networks (GANs), employing iterative point tracking and motion supervision in latent space. With the rise of diffusion models, a series of diffusion-based methods have also been proposed~\cite{shi2024dragdiffusion,hou2024easydrag,choi2024dragtext,zhang2024gooddrag}. For instance, DragDiffusion~\cite{shi2024dragdiffusion} also achieves drag-based image editing through point tracking and motion supervision. On this basis, EasyDrag~\cite{hou2024easydrag} and DragText~\cite{choi2024dragtext} have made significant improvements. There are also some other methods that utilize feature correspondences to achieve point-based image editing.
For example, DragonDiffusion~\cite{mou2023dragondiffusion} and DragEditor~\cite{mou2024diffeditor} utilizes an energy function to constructs gradient guidance that enable drag instructions to gradient-based process. However, they face challenges in balancing point operations and require more processing time.

\paragraph{One-step Optimization.} Although there are one-step generation works like InstaFlow ~\cite{liu2024instaflowstephighqualitydiffusionbased} in the field of text-to-image generation, FastDrag ~\cite{zhao2024fastdrag}, which edits  images based on geometric relationships, is the first work in the point-based image editing task. This approach significantly reduces time overhead compared to multi-step optimization. However, it still encounters challenges related to semantic inconsistency.

\begin{figure*}[]
    \centering
    \includegraphics[width=1.0\textwidth]{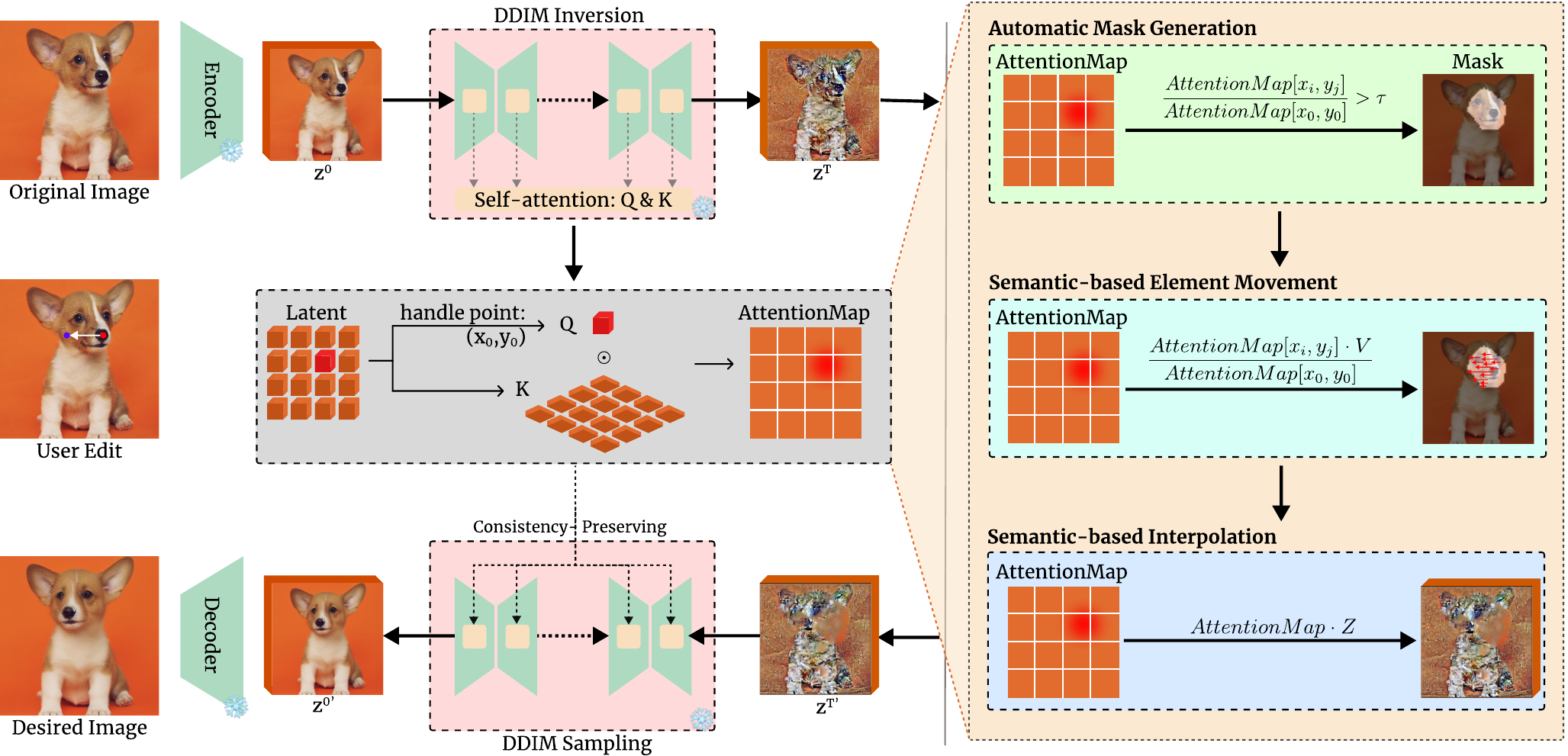}
    \caption{The pipeline of our framework, which mainly consists of Automatic Mask Generation, Semantic-based Element Movement and Semantic-based Interpolation.}
    \label{fig:model}
\end{figure*}

\section{Methodology}
As shown in Fig \ref{fig:model}, AttentionDrag is an attention-driven framework which leverages the self-attention mechanism within the U-Net model to enhance drag-based image editing through three phases: (1) Semantic-based Element Movement to ensure precise element relocation, (2) Automatic Mask Generation by analyzing semantic relationships between the handle point and other image elements, and (3) Semantic-based Interpolation using attention-guided interpolation to preserve semantic consistency.

\subsection{Preliminaries}
DDIM is an improvement over traditional diffusion models that accelerates the generation process by introducing an implicit generative process. This is achieved by modeling the reverse process using a non-Markovian process instead of a traditional Markov process, making the generation process more efficient.

\paragraph{Forward Diffusion Process.} In the forward process, a data sample \( z_0 \) (e.g., an image) is gradually corrupted by noise, forming a sequence \( z_1, z_2, \dots, z_T \), where \( T \) is the number of time steps. Each step of the forward process is described by the following equation:
\begin{equation}
q(z_t | z_{t-1}) = \mathcal{N}(z_t; \sqrt{1 - \beta_t} z_{t-1}, \beta_t \mathds{I}),
\end{equation}
Here, \( \beta_t \) is the noise variance at each time step, typically increasing over time.

\paragraph{Reverse Diffusion Process.} The reverse process aims to recover the original sample \( z_0 \) from the noise \( z_T \). This is done by learning the reverse conditional probabilities, modeled as:
\begin{equation}
p_{\theta}(z_{t-1} | z_t) = \mathcal{N}(z_{t-1}; \mu_{\theta}(z_t, t), \Sigma_t),
\end{equation}
where, \( \mu_{\theta}(z_t, t) \) is the mean predicted by the model, and \( \Sigma_t \) is the covariance term.

\paragraph{Stable Diffusion.} Stable Diffusion is a generative model based on latent diffusion that efficiently generates high-quality images by operating in the latent space. The use of the latent space significantly reduces computational complexity, enabling applications like text-to-image generation and image inpainting efficiently.

\subsection{Semantic-based Element Movement}
\label{section:Semantic-Based Element Movement}

\paragraph{Extracting the Attention Map.} During the DDIM forward process, the input image \(z_0\) is gradually corrupted into a noisy representation \(z_T\) over \(T\) timesteps. At each timestep \(t\), the noisy latent \(z_t\) is processed through the U-Net to predict noise components \( \epsilon_\theta(z_t, t) \). While predicting \( \epsilon_\theta(z_t, t) \), the U-Net computes self-attention scores that quantify the semantic relationships between all spatial positions in the latent representation. The attention scores at timestep \(t\) are calculated by:

\begin{equation}
A_t[x_0, y_0, x_i, y_j] = \frac{\exp\left(\mathbf{q}[x_0, y_0] \cdot \mathbf{k}[x_i, y_j] / \sqrt{d_k}\right)}{\sum_{x, y} \exp\left(\mathbf{q}[x_0, y_0] \cdot \mathbf{k}[x, y] / \sqrt{d_k}\right)},
\end{equation}
where, \( \mathbf{q}[x_0, y_0] \) and \( \mathbf{k}[x_i, y_j] \) are the query and key vectors at handle point \((x_0, y_0)\) and other point \((x_i, y_j)\), respectively. And \( d_k \) is the dimensionality of the key vectors.

The attention map is extracted from the 4D self-attention score tensor \( A_t \) by slicing along the dimensions corresponding to the handle point \( (x_0, y_0) \). Therefore, \(AttentionMap_t\) at timestep \(t\) can be formulated as:

\begin{equation}
\begin{aligned}
    AttentionMap_t &= A_t[x_0, y_0, :, :], \\
    AttentionMap_t[x_i, y_j] &= A_t[x_0, y_0, x_i, y_j], 
\end{aligned}
\end{equation}
where, the value \(AttentionMap_t[x_i, y_j]\) represents the semantic relevance between the handle point \((x_0, y_0)\) and other point \((x_i, y_j)\).

\paragraph{Calculating Movement Vectors.} The displacement of each position \((x_i, y_j)\) in the latent space is governed by the corresponding attention score and the user-provided drag instruction. The drag instruction specifies the target displacement of the handle point, defined as:

\begin{equation}
V = (x_{\text{target}} - x_0, y_{\text{target}} - y_0) = (\Delta x_0, \Delta y_0),
\end{equation}
where \((x_{\text{target}}, y_{\text{target}})\) is the target position for the handle point \((x_0, y_0)\). By using the attention map, the movement vector \(v_t[x_i, y_j]\) at timestep \(t\) for position \((x_i, y_j)\) is calculated as:

\begin{equation}
v_t[x_i, y_j] = \frac{AttentionMap_t[x_i, y_j] \cdot V}{AttentionMap_t[x_0, y_0]} = (\Delta x_i, \Delta y_j),
\end{equation}
where \(AttentionMap_t[x_0, y_0]\) serves as a reference value, ensuring stability and adaptability across different images. This formulation ensures that the displacement of each position is proportional to its semantic correlation with the handle point.

To enhance robustness and ensure semantic consistency, the movement vectors are aggregated across multiple timesteps during the DDIM inversion process. The final latent representation is updated as:

\begin{equation}
v[x_i, y_j] = \frac{1}{N} \sum_{t \in N} v_t[x_i, y_j],
\end{equation}
where, \(N\) is the number of inversion steps in diffusion inversion.

Then, the movement vector is limited by the generated mask \(\textbf{M}(x_i, y_j)\) (will be discussed in Sec. \ref{section:Automatic Mask Generation}), ensuring that adjustments occur exclusively within the mask region. The revised movement vector is given by:

\begin{equation}
\begin{aligned}
v[x_i, y_j] &= \textbf{M}(x_i, y_j) \cdot v[x_i, y_j] \\
&=
\begin{cases} 
(\Delta x_i, \Delta y_j), &\text{if } \textbf{M}(x_i, y_j) == 1, \\
(0, 0), &\text{otherwise}.
\end{cases}
\end{aligned}
\end{equation}
where, \(\textbf{M}(x_i, y_j) \in \{0, 1\}\) is the mask value, where 1 indicates the position is within the editing region, and 0 indicates it is outside.

\paragraph{Updating the Latent Representation.} The latent representation \(z_T\) is updated by applying the computed movement vectors to each position \((x_i, y_j)\) where \(\textbf{M}(x_i, y_j) == 1\):

\begin{equation}
z'[x_i+\Delta x_i, y_j+\Delta y_j] = z[x_i, y_j],
\end{equation}
where, \(z[x_i, y_j]\) is the original latent value at position \((x_i, y_j)\), \(z'\) is the updated latent representation.

\subsection{Automatic Mask Generation}
\label{section:Automatic Mask Generation}
\paragraph{Extracting the Attention Map.} As discussed in Sec. \ref{section:Semantic-Based Element Movement}, the attention map \(AttentionMap_t[x_i, y_j]\) at timestep \(t\) is extracted from the 4D attention tensor \(A_t[x_0, y_0, x_i, y_j]\):

\begin{equation}
AttentionMap_t[x_i, y_j] = A_t[x_0, y_0, x_i, y_j].
\end{equation}

To ensure the stability of the attention map, which is crucial for generating adaptive masks, we aggregate attention maps across multiple timesteps during the DDIM inversion process. Specifically, for each timestep \( t \), the attention map \(AttentionMap_t[x_i, y_j]\) is extracted as described earlier. These maps are then aggregated over a set of timesteps \(N\) to form a combined attention map:

\begin{equation}
AttentionMap[x_i, y_j] = \frac{1}{N} \sum_{t \in N} AttentionMap_t[x_i, y_j],
\end{equation}
where, \(AttentionMap_t[x_i, y_j]\) is the attention map at timestep \(t\), \(N\) is the set of timesteps used for aggregation.

\paragraph{Generating the Binary Mask.} Using the extracted attention map, a binary mask \textbf{M} is generated by applying a predefined threshold \(\tau\). The mask is defined as:

\begin{equation}
\textbf{M}(x_i, y_j) =
\begin{cases} 
1, & \text{if } \frac{AttentionMap[x_i, y_j]}{AttentionMap[x_0, y_0]} > \tau, \\
0, & \text{otherwise}.
\end{cases}
\end{equation}
Here, \(\textbf{M}(x_i, y_j) = 1\) indicates that the position \((x_i, y_j)\) is included in the editing region, \(\tau\) controls the sensitivity of the mask generation.

\subsection{Semantic-based Interpolation}
In the semantic-based element movement operation, moving elements will create blank regions. When the latent value \( z[x_i, y_j] \) at \( (x_i, y_j) \) is moved, the original position becomes blank. To preserve semantic consistency and visual coherence, we propose an attention-based interpolation strategy that uses self-attention scores in the U-Net block during DDIM Inversion to capture semantic relationships between blank and surrounding regions, reconstructing the blanks by aggregating surrounding information.

For each blank region \((x_i, y_j)\), its attention score with respect to other region \((x_n, y_m)\) is \(AttentionMap[x_n, y_m]\), which is extracted from the self-attention mechanism in the U-Net module during the DDIM inversion process, as described in Sec. \ref{section:Semantic-Based Element Movement}.

This attention map is then used to guide the interpolation process, where the values of the blank regions are reconstructed based on the semantic correlations with surrounding areas.

\begin{equation}
\hat{z}[x_i, y_j] = \sum_{x_n, y_m} AttentionMap[x_n, y_m] \cdot z'[x_n, y_m],
\end{equation}
where \(\hat{z}[x_i, y_j]\) is the interpolated value at blank region \((x_i, y_j)\).

\begin{figure*}[t]
    \flushright
    \includegraphics[width=0.99\textwidth]{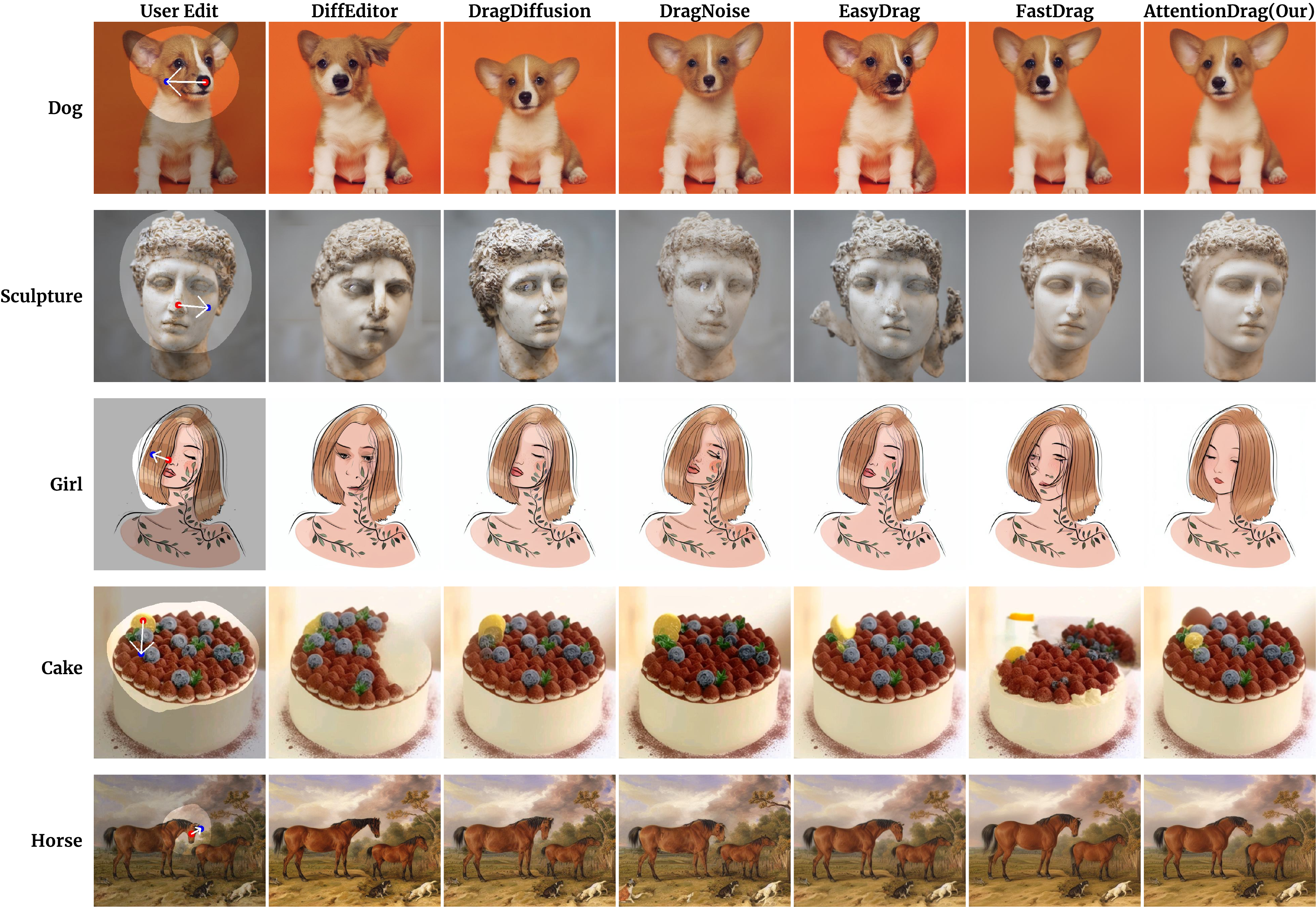}
    \caption{Illustration of image drag editing performance of single-point manipulation compared to current state-of-the-art methods.}
    \label{fig:single}
\end{figure*}

\section{Experiments}

\subsection{Implementation Details}
We employ a pre-trained Latent Diffusion Model (LDM), specifically Stable Diffusion 1.5 ~\cite{manukyan2023hd}, as the underlying diffusion model. The U-Net architecture is adapted with LCM-distilled weights from Stable Diffusion 1.5. By default, the inversion and sampling steps are set to 10, unless stated otherwise. Following DragDiffusion ~\cite{shi2024dragdiffusion}, we do not apply classifier-free guidance (CFG) ~\cite{ho2022classifier} in our diffusion model, and the latent diffusion optimization is performed at the 5th step. All other configurations are aligned with those used in DragDiffusion~\cite{shi2024dragdiffusion}. The experiments are conducted on an H20 GPU with 96GB of memory.

\subsection{Qualitative Evaluation}
\begin{figure*}[t]
    \centering
    \includegraphics[width=1.0\textwidth]{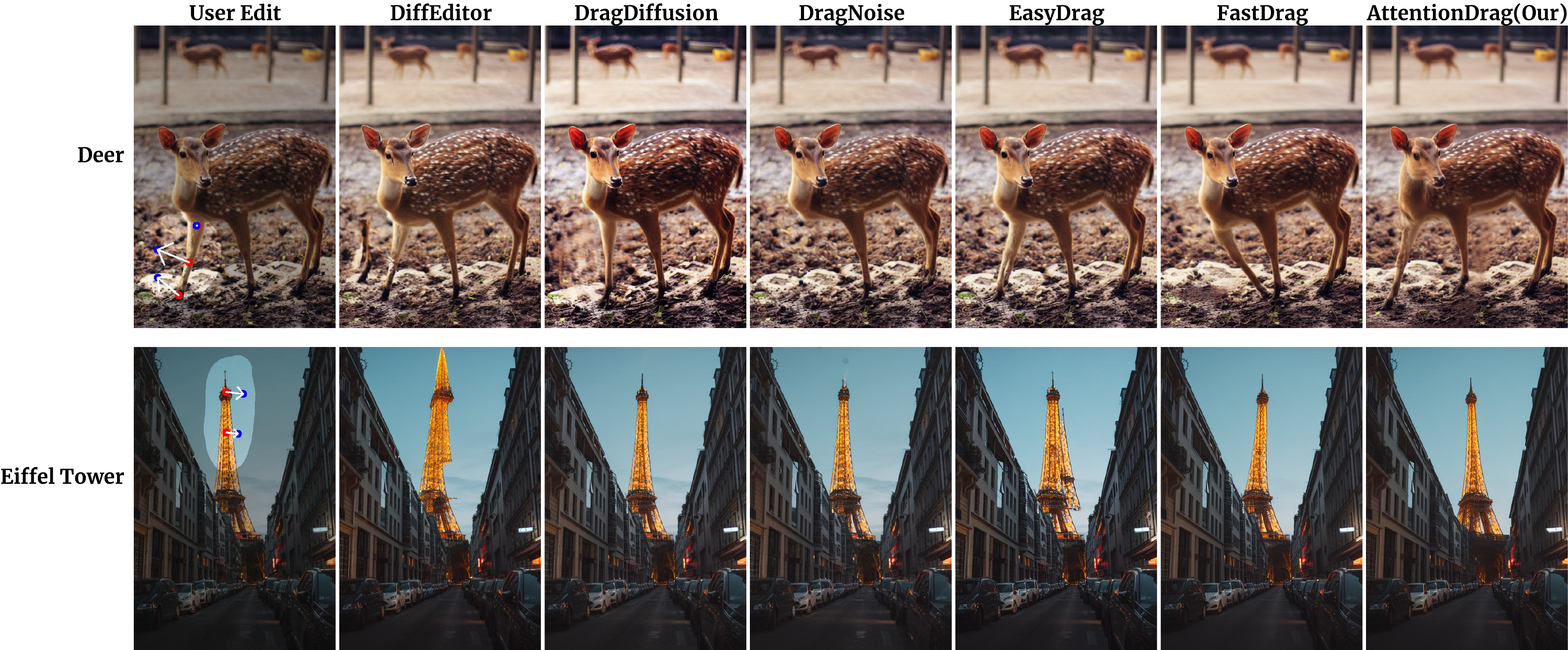}
    \caption{Illustration of image drag editing performance of multi-point manipulation compared to current state-of-the-art methods.}
    \label{fig:multi}
\end{figure*}

\paragraph{Comparative Results.}As shown in Fig. \ref{fig:single} and Fig. \ref{fig:multi}, we provide a comparative analysis of the performance of AttentionDrag in point-based image editing tasks. The results are shown in two sets of images: (a) for single-point manipulation and (b) for multi-point manipulation. In these comparisons, we evaluate AttentionDrag alongside several state-of-the-art point-based image editing methods, including DiffEditor, DragDiffusion, DragNoise, EasyDrag, and FastDrag.

In single-point image editing tasks, AttentionDrag demonstrates exceptional performance by precisely modifying the target region without disrupting the overall consistency of the image. For example, in the Dog image, AttentionDrag successfully alters the shape of the ears while maintaining a natural transition with the surrounding background, avoiding the distortions and inconsistencies often seen in methods like DiffEditor and FastDrag. Similarly, in the Sculpture and Girl images, AttentionDrag effectively preserves fine details and textures, ensuring visual coherence between the edited and unedited parts of the image. Note that due to a code issue with EasyDrag, the image of the edited horse comes from the result after removing the blue dot from the image in the paper.

In multi-point image editing tasks, AttentionDrag effectively handles large-scale object movements while maintaining spatial consistency and semantic coherence across the image. For instance, in the Deer image, AttentionDrag successfully moves the deer without introducing misalignment or artifacts in the background, unlike DragNoise and DiffEditor, where transitions between modified and unmodified areas appear unnatural. In the Eiffel Tower image, AttentionDrag ensures smooth transitions between the foreground and background, avoiding the misalignments seen in EasyDrag, and maintaining a harmonious visual result.

\begin{figure}[H]
    \centering
    \includegraphics[width=1\linewidth]{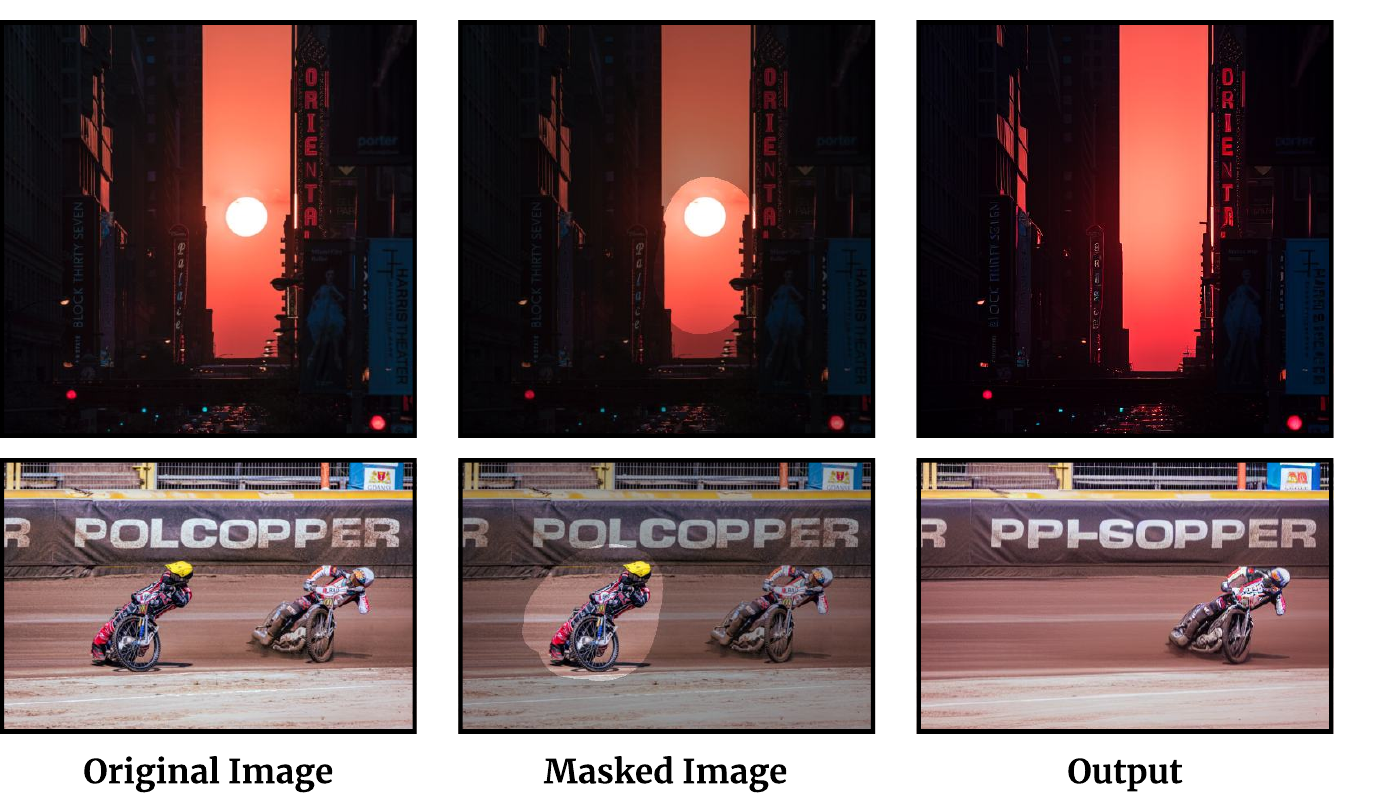}
    \caption{Qualitative results for image inpainting.}
    \label{fig:inpaint}
\end{figure}

\paragraph{Inpainting Results.} Interestingly, our method is not only applicable to point-based image editing tasks. When drag point information is not provided and only mask information is available, our proposed approach can even perform effectively in inpainting tasks. 
As demonstrated in Fig. \ref{fig:inpaint}, the masked areas are successfully filled with semantically consistent and visually coherent content, seamlessly blending with the surrounding regions. It is noteworthy that this method is plug-and-play with any diffusion model framework, allowing for easy adaptation to various low-level visual tasks.

In summary, AttentionDrag is highly effective in preserving semantic consistency, accurately handling object movement, and retaining fine details in both simple and complexscenes.

\begin{table}[]
\centering
\renewcommand{\arraystretch}{1.1}
\resizebox{\linewidth}{!}{
\begin{tabular}{cccccc}
\hline
\hline
\multirow{2}{*}{Approach} & \multirow{2}{*}{MD $\downarrow$} & \multirow{2}{*}{1-LPIPS $\uparrow$} & \multicolumn{2}{c}{Time}  \\ \cline{4-5} 
                          &                             &                           & Preparation  & Editing(s) \\ \hline
DragDiffusion             & 33.70               & \textcolor{red}{0.89}                     & 1 min (LORA) & 21.54      \\
DragNoise                 & 33.41               & 0.63                     & 1 min (LORA) & \underline{20.41}      \\
FreeDrag                  & 35.00               & 0.70                     & 1 min (LORA) & 52.63      \\
GoodDrag                  & \textcolor{red}{22.96}               & \underline{0.86}                     & 1 min (LORA) & 45.83      \\
DiffEditor                & \textcolor{blue}{28.46}               & \textcolor{red}{0.89}                     &   \XSolidBrush        & 21.68      \\ \hline
FastDrag                  & 32.23               & \underline{0.86}                      &    \XSolidBrush          & \textcolor{blue}{5.66}       \\
\textbf{Ours}       & \textbf{\underline{29.66}}             & \textbf{\textcolor{blue}{0.87}}                  &         \XSolidBrush     & \textbf{\textcolor{red}{5.50}}  \\ \hline
\hline
\end{tabular}}
\caption{Performance comparison of different approaches. MD and 1-LPIPS indicate metric values, while time preparation and editing time are provided for each method.}
\label{tab:result}
\end{table}

\subsection{Quantitative Analysis}

In order to rigorously evaluate the performance of \textbf{AttentionDrag}, we conduct a quantitative comparison using the DragBench dataset, which consists of 211 distinct image categories with 349 handle-target point pairs. Following FastDrag~\cite{zhao2024fastdrag}, the evaluation is based on two key metrics: mean distance (MD) ~\cite{pan2023drag} and image fidelity (IF) ~\cite{kawar2023imagic}. The MD metric quantifies the precision of the drag operation by measuring the displacement of the edited point, while the IF metric assesses the perceptual similarity between the generated and original images, using the learned perceptual image patch similarity (LPIPS) ~\cite{zhang2018unreasonable}. Specifically, we utilize 1-LPIPS as the IF measure to facilitate direct comparison. In addition, we report the average time taken per point to highlight the computational efficiency of AttentionDrag. 

We selected several of the most recent point-based image editing methods for comparison. As shown in the Table \ref{tab:result}, AttentionDrag, our proposed one-step, training-free method, achieves a significant reduction in Mean Distance (MD) compared to FastDrag, another one-step method, while maintaining a similar 1-LPIPS score. This demonstrates that AttentionDrag, which leverages semantic correlations for image editing, outperforms FastDrag, which is based on geometry-based transformations. This advantage reflects the superior ability of semantic-based editing to preserve image coherence and semantic integrity.

In comparison to latent iterative optimization-based approaches, AttentionDrag not only reaches state-of-the-art (SOTA) performance in both LPIPS and MD, but also significantly reduces the time required for editing. And the reduction in computational cost highlights the efficiency of our approach, establishing it as a robust and fast alternative to existing iterative methods. These results underscore the effectiveness of exploiting latent feature correlations in pre-trained diffusion models for high-quality, time-efficient image editing.

\subsection{Ablation Study}

\begin{figure}[t]
    \centering
    \includegraphics[width=1\linewidth]{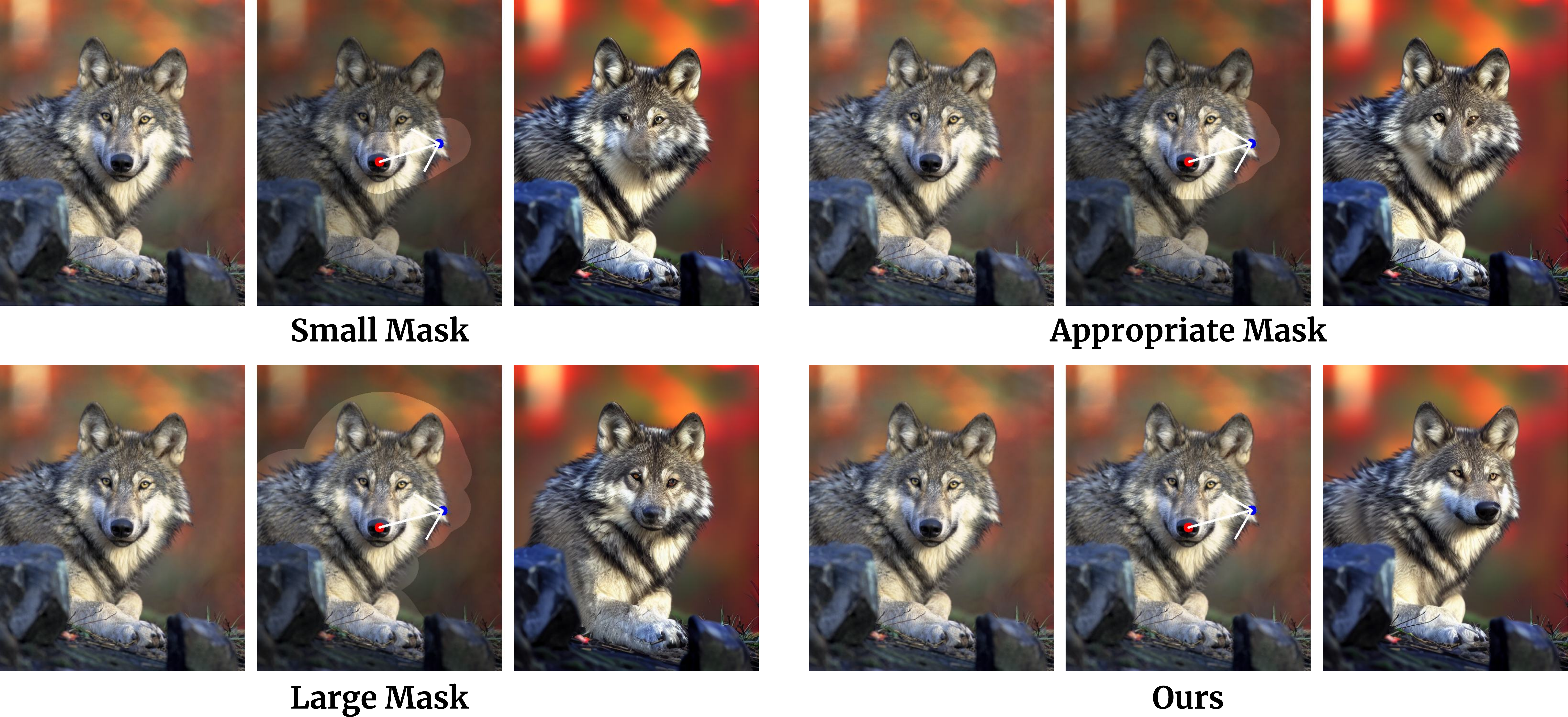}
    \caption{Ablation study on Automatic Mask Generation.}
    \label{fig:mask}
\end{figure}

\begin{figure}[t]
    \centering
    \includegraphics[width=1\linewidth]{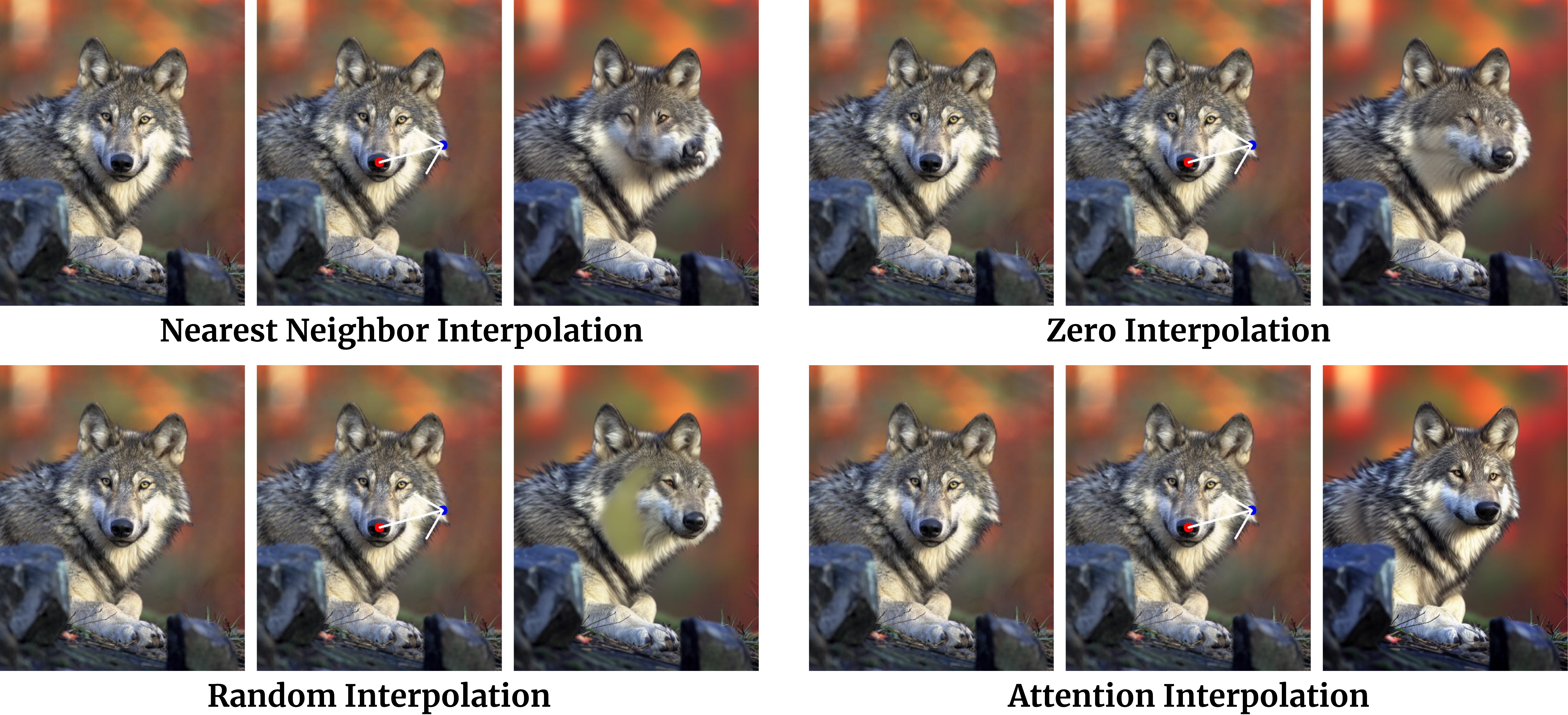}
    \caption{Ablation study on Semantic-based Interpolation.}
    \label{fig:interpolation}
\end{figure}

\begin{table}[t]
\centering
\setlength{\tabcolsep}{12pt}
\renewcommand{\arraystretch}{1.1}
\begin{tabular}{rcc}
\hline \hline
Model              & MD $\downarrow$   & 1-LPIPS $\uparrow$ \\ \hline
baseline             & 37.67 & 0.85    \\ 
w/o AMG         & 36.5  & 0.86    \\ 
w/o SBI & 32.0  & 0.85    \\ \hline
\textbf{ours}             & \textbf{\textcolor{red}{29.66}} & \textbf{\textcolor{red}{0.87}}    \\ \hline \hline
\end{tabular}
\caption{Ablation study of Automatic Mask Generation (w/o AMG) and Semantic-based Interpolation (w/o SBI).}
\label{tab:amg&si}
\end{table}

\paragraph{Automatic Mask Generation.} To demonstrate the effectiveness of the Automatic Mask Generation, we compare it with the results generated using several manually input masks. As shown in Fig \ref{fig:mask}, Automatic Mask Generation achieves the best performance in image editing tasks. Compared to small and large masks, which either under-represent or over-represent the editing region, our method generates an appropriately sized mask that precisely targets the area to be modified. As indicated in the Table \ref{tab:amg&si}, removing the automatic mask generation results in a slight increase in 1-LPIPS to 0.86, but the precision of the edits (MD) is reduced to 36.5, indicating that the mask is crucial for precise editing.


\paragraph{Semantic-based Interpolation.} Semantic-based Interpolation outperforms other interpolation methods in reconstructing blank spaces, as shown in Fig \ref{fig:interpolation}. While methods like Nearest Neighbor Interpolation, 0 Interpolation, and Random Interpolation introduce visible artifacts or unnatural transitions, Semantic-based Interpolation effectively leverages the surrounding context to fill blank areas seamlessly. By utilizing attention mechanisms, it ensures smooth integration of the edited region with the rest of the image, maintaining both visual coherence and semantic consistency, making it the most effective approach for blank space reconstructing. As presented in the Table \ref{tab:amg&si}, Omitting semantic-based interpolation improves the Mean Distance (MD) to 32.0, but the 1-LPIPS decreases to 0.85, suggesting a trade-off between spatial accuracy and semantic consistency.

\begin{figure}[H]
    \centering
    \includegraphics[width=0.9\linewidth]{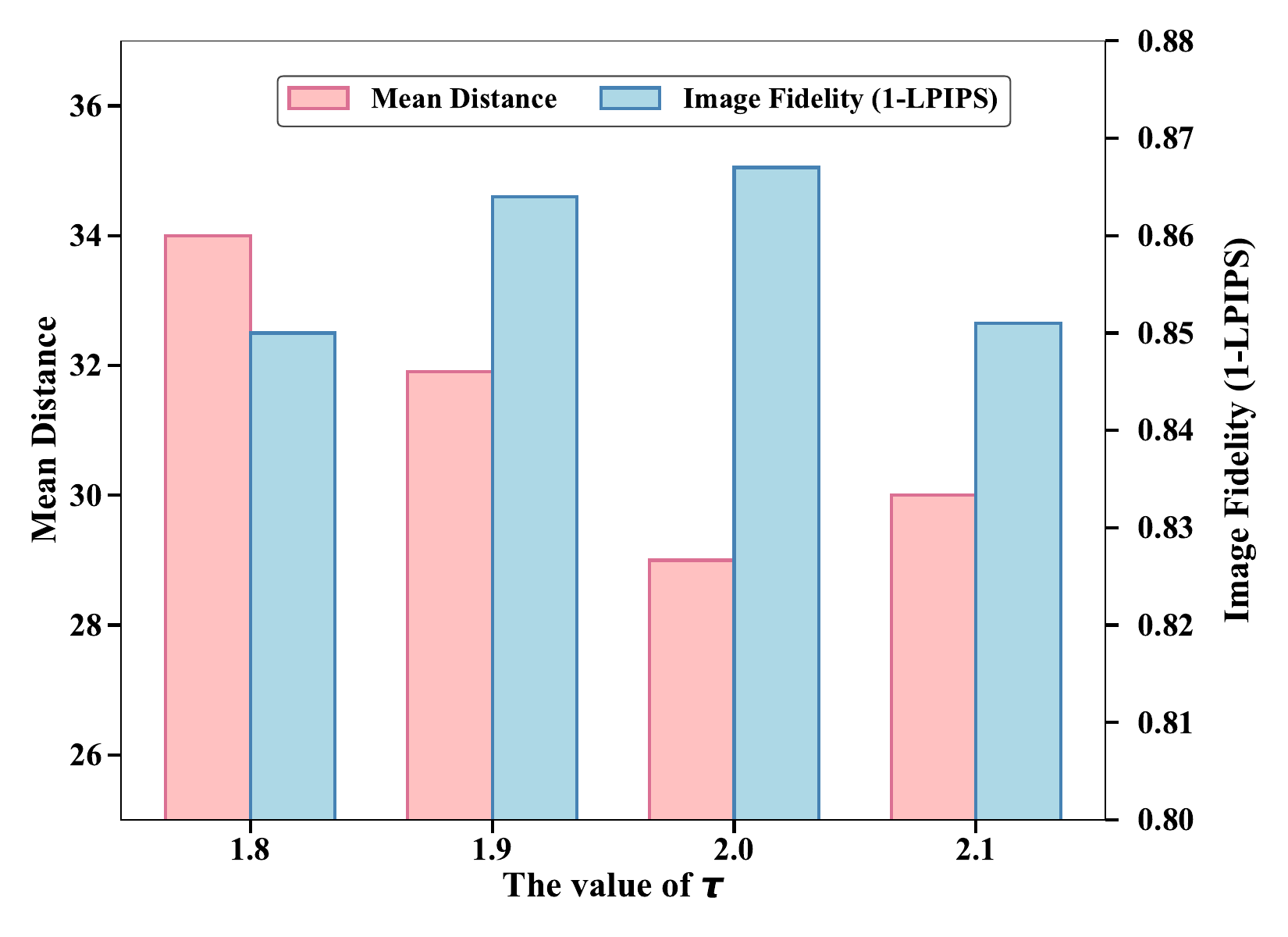}
    \caption{Ablation study on the value of \(\tau\).}
    \label{fig:ab}
\end{figure}

\paragraph{The value of \(\tau\).} Finally, we conduct an ablation study to elucidate the impact of varying \(\tau\). We set \(\tau\) to be \( \tan = 1.8,1.9,2.0,2.1\) and run our approach on DragBench dataset to get the editing results. The outcomes are as- sessed with IF and MD, and are shown in Fig. \ref{fig:ab}. As \(\tau\) increases from 1.8 to 2.0, the Mean Distance decreases and then increases slightly at \(\tau\) = 2.1, indicating an optimal balance at \(\tau\) = 2.0. Meanwhile, Image Fidelity improves with \(\tau\), reaching its peak at 2.1, reflecting worse semantic preservation. This study highlights the importance of tuning \(\tau\) to achieve a balance between precise manipulation and high semantic consistency.

\section{Conclusion and Future Works}
In this paper, we introduced AttentionDrag, a novel, one-step, training-free method for point-based image editing that effectively harnesses the latent knowledge and feature correlations from pre-trained diffusion models. Our method outperforms existing approaches by preserving semantic consistency and producing high-quality edits with minimal computational overhead. Future work could focus on extending AttentionDrag to more complex and diverse datasets, enhancing real-time performance, and exploring its potential for interactive, user-guided image editing applications.

\appendix



\section*{Acknowledgments}
This work was supported by the University-Industry Collaborative Research Project between Fudan University and Ele.me under Alibaba Group.

\bibliographystyle{named}
\bibliography{ijcai25}

\end{document}